# Graph-Based Deep Learning for Intelligent Detection of Energy Losses, Theft, and Operational Inefficiencies in Oil & Gas Production Networks

*AbdulQoyum A. Olowookere[1], Adewale U. Oguntola[2], Ebenezer Leke Odekanle[3]

[1,3]Department of Chemical and Petroleum Engineering, Abiola Ajimobi Technical University, Ibadan, Oyo State, Nigeria.

[1*]abdulqoyumadegoke@gmail.com
[3]odekanleebenezer@gmail.com

[2]Department of Computer Science, Abiola Ajimobi Technical University, Ibadan, Oyo State, Nigeria
[2]uoguntola@gmail.com

* AbdulQoyum A. Olowookere

**Abstract.** Early detection of energy losses, theft, and operational inefficiencies remains a critical challenge in oil and gas production systems due to complex interdependencies among wells and facilities, evolving operating conditions, and limited labeled anomaly data. Traditional machine learning approaches often treat production units independently and struggle under temporal distribution shifts. This study proposes a spatiotemporal graph-based deep learning framework for anomaly detection in oil and gas production networks. The production system is modeled as a hierarchical graph of wells, facilities, and fields, with additional peer connections among wells sharing common infrastructure. Weakly supervised anomaly labels are derived from physically informed heuristics based on production, pressure, and flow behavior. Temporal dynamics are captured through sequence modeling, while relational dependencies are learned using a Temporal Graph Attention Network. Under time-based evaluation, the proposed model achieves an ROC-AUC of about 0.98 and anomaly recall above 0.93, demonstrating improved robustness and practical potential for proactive monitoring in real-world energy operations.

# 1    Introduction

The detection of anomalies in oil and gas production systems is a long-standing challenge with significant implications for operational efficiency, economic performance, and environmental safety. Anomalies such as unexpected production declines, abnormal pressure–flow relationships, equipment malfunction, and unauthorized losses or theft can lead to substantial revenue loss if not identified early. Modern production environments generate large volumes of time-varying data from wells, sensors, and surface facilities, making manual monitoring and rule-based detection increasingly inadequate.

Anomaly detection has been extensively studied within the broader machine learning literature, where anomalies are generally defined as patterns in data that deviate significantly from expected behavior (Chandola et al., 2009). Traditional approaches include statistical process control, threshold-based rules, and distance-based techniques; however, these methods often struggle in high-dimensional, non-stationary environments typical of industrial systems. More recent machine learning-based approaches, including supervised and semi-supervised learning, have demonstrated improved performance by learning complex nonlinear relationships from historical data (Ahmed et al., 2016). Nevertheless, many of these methods assume independent samples and fail to account for temporal dynamics or structural dependencies among system components.

In the oil and gas domain, machine learning has increasingly been applied to anomaly detection tasks such as pipeline leak detection, production monitoring, and asset integrity assessment. Supervised learning models, including random forests, support vector machines, and ensemble methods, have been shown to outperform classical techniques in identifying abnormal operating conditions in pipeline systems and production facilities (Wu et al., 2019; Adewole et al., 2022; Aljameel et al., 2022). Deep learning models have been explored for pipeline health assessment and leakage detection, particularly under imbalanced data conditions (Shaik et al., 2024; Yang et al., 2025). Despite these advances, most existing approaches treat pipelines, wells, or sensors as isolated entities, neglecting the fact that oil and gas production systems operate as interconnected networks in which wells, manifolds, separators, and facilities are physically and operationally linked.

Production networks inherently exhibit relational structure: multiple wells feed into shared facilities, production behavior at one node can influence others, and failures or inefficiencies may propagate across the network. Graph-based representations provide a natural framework for modeling such systems, where nodes represent production entities and edges encode physical or functional dependencies. Graph Neural Networks (GNNs) have emerged as a powerful class of deep learning models capable of learning from graph-structured data by aggregating information from neighboring nodes (Kipf and Welling, 2017). Extensions such as Graph Attention Networks further improve representational capacity by assigning adaptive importance weights to neighboring nodes during message passing (Veličković et al., 2018).

Recent studies have demonstrated the effectiveness of GNNs in energy and industrial systems, including smart grids, pipeline networks, and reservoir management applications. In the power domain, GNN-based models have been successfully applied to electricity theft detection and energy anomaly identification, outperforming traditional deep learning approaches by exploiting network topology (Liao et al., 2023; Malik et al., 2024). In the oil and gas context, graph-based learning has been explored for tasks such as gas pipeline property estimation and production optimization, highlighting the ability of GNNs to capture complex physical dependencies in energy infrastructure (Yang et al., 2023; Li et al., 2025). Furthermore, recent work has shown that combining graph learning with temporal attention mechanisms enables effective detection of collective and contextual anomalies in multivariate industrial time series (Zhao et al., 2020; Deng and Hooi, 2021).

In addition to spatial dependencies, oil and gas production data exhibit strong temporal characteristics driven by reservoir dynamics, operational interventions, and equipment aging. Sequence-based deep learning models such as Long Short-Term Memory (LSTM) networks have been widely used to model time-series data and detect anomalies based on evolving temporal patterns (Malhotra et al., 2015; Hundman et al., 2018). More recent spatiotemporal deep learning models have demonstrated improved performance in oilfield production forecasting by jointly modeling temporal trends and spatial interactions among wells (Du et al., 2023; Hu et al., 2024; Lu et al., 2025). While effective, purely temporal models remain limited in their ability to exploit explicit production network structure for anomaly detection.

Motivated by these limitations, this study proposes a graph-based deep learning framework for intelligent detection of energy losses, theft, and operational inefficiencies in oil and gas production networks. The proposed approach models the production system as a hierarchical graph comprising wells, facilities, and fields, and integrates temporal modeling with graph attention mechanisms to capture both sequential dynamics and relational dependencies. Weakly supervised labels derived from physically meaningful heuristics are used to enable scalable training in the absence of comprehensive ground-truth annotations. The framework is evaluated using time-based splits to reflect real-world deployment scenarios, and its performance is compared against traditional machine learning and sequence-only deep learning baselines. The results demonstrate that incorporating graph structure improves anomaly detection performance, particularly in terms of recall and robustness under future data shifts.

The remainder of this paper is organized as follows. Section 2 reviews related work on anomaly detection, machine learning applications in energy systems, and graph-based deep learning approaches relevant to oil and gas production networks. Section 3 presents the proposed methodology, including the dataset description, feature engineering, weakly supervised labeling strategy, graph construction, and the Temporal Graph Attention Network architecture. Section 4 describes the experimental setup and discusses the results obtained from both baseline models and the proposed framework, including ablation analysis and evaluation under time-based data splits. Finally, Section 5 summarizes the main findings of the study and outlines directions for future research.

## 2     Related Work

Research on anomaly detection in industrial and energy systems has evolved significantly over the past two decades, driven by the increasing availability of high-frequency sensor data and the growing complexity of production infrastructure. In the oil and gas sector, anomaly detection is particularly important due to the high economic value of production assets and the potential environmental consequences of undetected failures or losses. Existing studies can broadly be categorized into three groups: (i) traditional statistical and machine learning approaches, (ii) deep learning models for temporal anomaly detection, and (iii) graph-based methods for modeling relational dependencies in industrial systems.

Early approaches to anomaly detection in oil and gas operations relied heavily on rule-based systems, control charts, and statistical thresholds derived from historical averages. While these methods are easy to implement, they are sensitive to noise and often fail under changing operating conditions (Chandola et al., 2009). As data availability increased, supervised machine learning techniques such as support vector machines, random forests, and gradient boosting gained popularity for tasks including pipeline leak detection, production monitoring, and equipment fault diagnosis.

Several studies have demonstrated the effectiveness of tree-based and kernel-based models in detecting abnormal events in oil and gas pipelines. For instance, Wu et al. (2019) applied ensemble learning techniques to identify pipeline leakage using pressure and flow data, achieving higher accuracy than conventional threshold-based methods. Similarly, Adewole et al. (2022) evaluated multiple machine learning algorithms for pipeline leak detection and showed that random forest and gradient boosting models achieved strong classification performance. Despite their success, these approaches typically assume independent observations and do not explicitly model temporal dependencies or interactions between different production units.

To address the limitations of static machine learning models, deep learning techniques have been increasingly adopted for time-series anomaly detection. Recurrent neural networks, particularly Long Short-Term Memory (LSTM) networks, are well suited for capturing long-term temporal dependencies in sequential data. Malhotra et al. (2015) demonstrated that LSTM-based models can effectively learn normal system behavior and detect deviations in multivariate time series. Subsequent studies extended this approach to industrial monitoring and energy systems, showing improved detection of gradual degradation and complex temporal patterns (Hundman et al., 2018).

In oil and gas applications, LSTM models have been used to analyze production trends, pressure variations, and equipment performance over time. While these models capture temporal dynamics more effectively than traditional methods, they generally treat each well or sensor stream independently. As a result, they may fail to detect anomalies that arise from interactions among multiple wells or shared facilities, which are common in production networks.

Graph-based learning has emerged as a powerful paradigm for modeling systems composed of interconnected entities. Graph Neural Networks (GNNs) extend deep learning to graph-structured data by propagating information along edges connecting related nodes (Kipf and Welling, 2017). Graph Attention Networks further enhance this framework by learning adaptive weights that quantify the importance of neighboring nodes during message passing (Veličković et al., 2018).

Recent research has explored the use of GNNs for anomaly detection in industrial and cyber-physical systems. Zhao et al. (2020) proposed a graph-based framework for detecting anomalies in multivariate time series by jointly modeling temporal patterns and graph structure. Deng and Hooi (2021) demonstrated that graph neural networks are particularly effective at detecting collective and contextual anomalies that cannot be identified through isolated analysis of individual nodes. In Industrial Internet of Things (IIoT) environments, GNN-based models have shown superior performance in capturing system-level abnormal behavior compared to traditional and sequence-only models.

Despite these advances, the application of graph-based deep learning to oil and gas production networks remains limited. Most existing studies focus on pipelines or individual sensors rather than hierarchical production systems involving wells, facilities, and fields. Furthermore, relatively few works integrate graph modeling with temporal deep learning to address the joint spatiotemporal nature of production anomalies.

From the reviewed literature, three key gaps can be identified. First, many machine learning approaches for oil and gas anomaly detection neglect the relational structure of production networks. Second, deep learning models that capture temporal dynamics often fail to incorporate spatial dependencies among interconnected assets. Third, limited attention has been given to realistic evaluation strategies, such as time-based splits, that reflect real-world deployment scenarios. This study addresses these gaps by proposing a graph-based temporal deep learning framework tailored to oil and gas production networks, enabling more robust detection of energy losses, theft, and operational inefficiencies.

| Author(s) | Study Purpose & Application Area | Methodology | Dataset | Technique | Key Findings | Limitations |
|---|---|---|---|---|---|---|
| Chandola et al. (2009) | General anomaly detection survey | Literature review | Multiple domains | Statistical & ML methods | Provided a foundational taxonomy of anomaly detection | Not application-specific |
| Wu et al. (2019) | Pipeline leak detection | Supervised learning | Pipeline pressure & flow data | Ensemble ML | Improved detection accuracy over threshold methods | Ignores temporal dependencies |
| Adewole et al. (2022) | Oil & gas pipeline monitoring | Comparative ML analysis | Simulated pipeline data | Random Forest, GBM | Tree-based models perform strongly | Treats pipelines as isolated systems |
| Malhotra et al. (2015) | Time-series anomaly detection | Deep learning | Multivariate time series | LSTM | Captures long-term temporal patterns | No relational modeling |
| Hundman et al. (2018) | Industrial system monitoring | Deep learning | Spacecraft telemetry | LSTM | Effective detection of gradual anomalies | Domain-specific |
| Kipf & Welling (2017) | Graph learning | Semi-supervised learning | Citation networks | Graph Convolutional Networks | Introduced a scalable GNN framework | Not focused on anomaly detection |
| Veličković et al. (2018) | Graph representation learning | Attention mechanisms | Benchmark graph datasets | Graph Attention Network | Improved node representations | No temporal modeling |
| Zhao et al. (2020) | Industrial anomaly detection | Graph-based modeling | Multivariate time series | Graph Neural Network | Captures system-level anomalies | Limited oil & gas focus |
| Deng & Hooi (2021) | Graph anomaly detection | Deep learning | Synthetic & real graphs | GNN-based models | Effective for collective anomalies | Evaluation mostly static |

| Author(s) | Study Purpose & Application Area | Methodology | Dataset | Technique | Key Findings | Limitations |
|---|---|---|---|---|---|---|
| This study | Energy loss, theft, and operational anomaly detection in oil & gas production networks | Spatiotemporal graph-based deep learning with weak supervision | Real-world oilfield production data (wells, facilities, fields) | Temporal Graph Attention Network (GAT) with hierarchical production graph | Improved anomaly recall (>0.93) and ROC-AUC (~0.98) under time-based evaluation by integrating temporal dynamics and relational dependencies | Evaluation conducted on a relatively small production network; weak labels derived from heuristic rules |

## 3. Methodology

This section describes the proposed spatiotemporal learning framework for detecting energy losses, theft, and operational inefficiencies in oil and gas production networks. The methodology integrates domain-informed feature engineering, weak supervision, graph-based representation of production infrastructure, and deep learning models that jointly capture temporal and relational dependencies.

Figure 1 illustrates the overall workflow of the proposed methodology. The process begins with production data collection and preprocessing, followed by feature engineering and weak anomaly label construction. The processed data are then transformed into temporal sequences and mapped onto a graph representation of the production network. Finally, the proposed Temporal Graph Attention Network learns both temporal dynamics and relational dependencies to detect anomalous operating conditions.

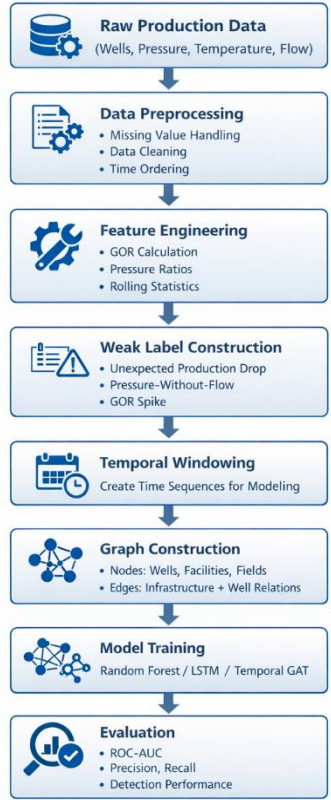

Figure 1: Overall methodological workflow for anomaly detection in oil and gas pre-production network

### 3.1 Problem Formulation

Oil and gas production systems comprise multiple interconnected wells that operate over time. For each well, multivariate operational data are recorded sequentially, reflecting changes in production behavior, pressure conditions, and interactions with the facility. The anomaly detection task is formulated as a supervised binary classification problem, where the objective is to predict whether a given production state corresponds to normal or anomalous operation.

Let $\mathcal{W} = \{w_1, w_2, \dots, w_N\}$ denote the set of production wells in the system. For each well $w_1$, observations are recorded at discrete time steps $t = 1, 2, \dots, T$. Each observation is represented by a feature vector $x_{i,t-r} \in \mathbb{R}^F$, where $F$ denotes the number of engineered features.

To enable early detection, predictions are based on historical information. Specifically, the model learns a mapping from a fixed-length temporal window of past observations to an anomaly label at the current time step. This relationship is formally defined as

$$f: \{x_{i,t-r:t-1}\} \to \widehat{y_{i,t}}, \qquad (1)$$

Where $r$ is the temporal window length and $\widehat{y_{i,t}} \in \{0,1\}$ denotes the predicted anomaly label for well $w_i$ at time $t$.

**3.2 Dataset Description and Preprocessing**

The dataset used in this study is derived from the Volve offshore field, a North Sea oil field operated historically under solution-gas drive and later subject to water breakthrough and artificial lift optimization. The dataset consists of real operational production records, including oil, gas, and water production volumes, pressure and temperature measurements, choke settings, and downtime indicators. Each record is indexed by production date and well identifier.

To ensure temporal consistency, the data were sorted chronologically on a per-well basis. Missing values were handled using engineering-aware strategies designed to preserve physical meaning. Production volumes were set to zero during confirmed non-producing periods, while short gaps in sensor measurements were forward-filled within limited horizons. For variables exhibiting persistent missingness, binary missingness indicators were introduced to retain information about sensor availability. This approach avoids aggressive imputation while allowing the model to distinguish between true operational signals and missing measurements.

**3.3 Feature Engineering**

Raw production and sensor measurements were augmented with temporal and physics-informed features to better capture operational dynamics. Temporal change rates were computed to quantify short-term fluctuations, as shown in Equation (2):

$$\Delta x_t = x_t - x_{t-1} \qquad (2)$$

Where $x_t$ represents the value of a given variable at time t.

To characterize local trends and variability, rolling statistics were computed over fixed windows of length $k$. The rolling mean and standard deviation are defined in Equations (3) and (4), respectively:

$$\mu_t^{(k)} = \frac{1}{k}\sum_{j=t-k+1}^{t} x_j \qquad (3)$$

$$\sigma_t^{(k)} = \sqrt{\frac{1}{k}\sum_{j=t-k+1}^{t}(x_j - \mu_t^{(k)})^2} \tag{4}$$

In addition, physics-based ratios commonly used in production engineering were included. The gas–oil ratio (GOR) and water cut were computed to capture abnormal production behavior, as shown in Equations (5) and (6):

$$GOR_t = \frac{Q_{g,t}}{Q_{o,t}+\epsilon'} \tag{5}$$

$$WaterCut_t = \frac{Q_{w,t}}{Q_{o,t}+Q_{w,t}+\epsilon'} \tag{6}$$

where $Q_{g,t}$, $Q_{o,t}$, and $Q_{w,t}$ denote gas, oil, and water production rates, respectively, and $\epsilon$ is a small constant added to ensure numerical stability.

### 3.4 Weak Supervision and Anomaly Label Construction

Due to the absence of comprehensive ground-truth annotations for energy losses and theft events, a weak supervision strategy was adopted. Anomalies were identified using a set of physically motivated heuristic rules designed to flag abnormal production behavior, such as unexpected production drops, abnormal pressure–flow relationships, and extreme GOR deviations.

Let $h_k(x_{i,t}) \in \{0,1\}$ denote the output of the $k$-th heuristic rule applied to the observation of the well $w_i$ at time $t$. The final weak anomaly label is assigned according to the logical aggregation shown in Equation (7):

$$y_{i,t} = \begin{cases} 1, & if\ \sum_k h_k(x_{i,t}) \geq 1, \\ 0, & otherwise \end{cases} \tag{7}$$

This formulation ensures that anomalies are flagged whenever at least one physically meaningful abnormal condition is detected.

### 3.5 Temporal Windowing

To model sequential dependencies while avoiding data leakage, a sliding temporal window approach was used. For each well $w_i$, an input sample consists of the previous $r$ observations leading up to time $t$. The temporal input representation is defined as

$$x_{i,t} = [x_{i,t-r}, x_{i,t-r+1}, \dots, x_{i,t-1}] \tag{8}$$

with the corresponding target label $y_{i,t}$. This causal formulation ensures that only historical information is used for prediction.

## 3.6 Graph Construction of the Production Network

Oil and gas production systems exhibit a strong relational structure, where multiple wells are connected to shared facilities and fields. To capture these dependencies, the production system is modeled as a directed graph $\mathcal{G} = (\mathcal{V}, \mathcal{E})$, where nodes represent wells, facilities, and fields, and edges encode physical and operational connections.

In addition to hierarchical connections, peer-to-peer edges are introduced between wells connected to the same facility. This design enables the model to capture shared operational context and correlated behavior across wells.

## 3.7 Temporal Graph Attention Network Architecture

Graph-based feature learning is performed using a Graph Attention Network (GAT), which assigns adaptive importance weights to neighboring nodes. The node embedding update at layer $l + 1$ is computed as shown in Equation (9):

$$\mathbf{z}_v^{(l+1)} = \sigma\left(\sum_{u \in \mathcal{N}(v)} \alpha_{vu}^{(l)} W^{(l)} z_u^{(l)}\right) \tag{9}$$

Where $\mathcal{N}(v)$ denotes the neighbors of the node $v$, $W^{(l)}$ is a learnable weight matrix, and $\sigma(\cdot)$ is a nonlinear activation function.

The attention coefficients $\alpha_{vu}^{(l)}$, which control the contribution of neighboring nodes, are computed using a softmax-normalized scoring function, as shown in Equation (10):

$$\alpha_{vu}^{(l)} = \frac{\exp(\text{LeakyRelu}(a^T[Wz_v]))}{\sum_{k \in \mathcal{N}(v)} \exp(\text{LeakyRelu}(a^T[Wz_v || Wz_k]))} \tag{10}$$

For each temporal window, the learned graph embedding of the corresponding well node is concatenated with the input time-series features and passed to an LSTM network, as shown in Equation (11):

$$h_{i,t} = LSTM(x_{i,t} \oplus \mathbf{z}_{w_i}) \tag{11}$$

Where $\oplus$ denotes feature concatenation. The final anomaly probability is obtained through a sigmoid activation function, as shown in Equation (12):

$$\widehat{y_{i,t}} = \sigma(w^T h_{i,t} + b) \tag{12}$$

## 3.8 Model Training and Evaluation

To address class imbalance inherent in anomaly detection problems, the model is trained using a weighted binary cross-entropy loss function. The loss formulation is shown in Equation (13):

$$\mathcal{L} = -[\beta y \log(\hat{y}) + (1 - y) \log(1 - \hat{y})], \quad (13)$$

Where $\beta$ is a weighting factor applied to the positive (anomalous) class. Model performance is evaluated using recall, precision, F1-score, and ROC-AUC, with particular emphasis on recall due to the high cost associated with missed anomaly events. Time-based train-test splits are employed to reflect realistic deployment scenarios.

## 4. Results and Discussion

This section presents and discusses the experimental results obtained from the proposed spatiotemporal graph-based anomaly detection framework. The analysis compares the proposed Temporal Graph Attention Network with baseline machine learning and deep learning models under both random and time-based evaluation settings. Quantitative metrics are complemented by visual analyses using ROC curves, Precision–Recall curves, and confusion matrices to provide a comprehensive assessment of model performance.

### 4.1 Performance under Random Train–Test Split

The random train–test split provides an initial benchmark for model performance under the assumption of stationary data distributions. Under this setting, all evaluated models achieved high overall accuracy and ROC-AUC values, reflecting the effectiveness of engineered features and weak supervision in separating normal and anomalous operating conditions.

The Random Forest model achieved the highest ROC-AUC of 0.984, indicating strong discriminative capability when trained on flattened tabular representations. However, its recall for anomalous events was limited to 0.794, suggesting that approximately 20% of anomaly windows were missed. This behavior highlights the limitation of static models in capturing evolving temporal patterns associated with production losses or inefficiencies.

In contrast, the LSTM model explicitly modeled temporal dependencies and improved anomaly recall to 0.824, demonstrating the importance of sequential information in anomaly detection tasks. Incorporating graph-based relational context through the Temporal Graph Attention Network further improved recall to 0.829, even though the underlying production network in this study was relatively small. The inclusion of additional well-well edges among wells connected to the same facility resulted in a modest but consistent improvement in ROC-AUC and F1-score, confirming the benefit of peer-level contextual information.

These results indicate that while traditional machine learning models perform well under random splits, temporal and relational modeling provides improved sensitivity to anomalous behavior.

This table reports ROC-AUC, precision, recall, and F1-score for the anomalous class. The Random Forest represents a traditional supervised baseline, while LSTM and Temporal GAT models incorporate temporal and spatiotemporal learning, respectively.

**Table 1. Performance of anomaly detection models under random train–test split**

| Model | ROC-AUC | Precision (Anomaly) | Recall (Anomaly) | F1-score |
|---|---|---|---|---|
| Random Forest | **0.984** | **0.951** | 0.794 | 0.865 |
| LSTM | 0.963 | 0.849 | 0.824 | 0.836 |
| Temporal GAT (hierarchy only) | 0.966 | 0.834 | 0.829 | 0.832 |
| Temporal GAT (with well–well edges) | 0.967 | 0.839 | 0.829 | 0.834 |

### 4.2 Performance under Time-Based Evaluation

To assess model robustness under realistic deployment conditions, a time-based split was applied, where models were trained on earlier production data and evaluated on future periods. This setting introduces a more challenging scenario due to distributional shifts and changes in operational conditions over time. Notably, the anomaly rate increased from approximately 3.8% in the training set to over 12% in the test set, emphasizing the importance of robust generalization.

Under this evaluation, deep learning models significantly outperformed traditional approaches. The LSTM achieved a ROC-AUC of 0.974 and an anomaly recall of

140.929, indicating strong sensitivity to future anomaly patterns. The proposed Temporal GAT further improved ROC-AUC to 0.980 and achieved the highest recall of 0.934, demonstrating superior discrimination and robustness under temporal distribution shift.

Although the Temporal GAT exhibited a slight reduction in precision compared to the LSTM (0.767 versus 0.784), this trade-off is acceptable in the context of energy loss and theft detection, where missed anomalies are substantially costlier than false positives. The improved recall achieved by the graph-based model underscores the value of incorporating relational context in time-evolving production systems.

This table compares the performance of temporal and spatiotemporal deep learning models under a realistic time-based evaluation. Metrics highlight the ability of each model to generalize to future production conditions.

**Table 2. Performance of deep learning models under time-based split**

| Model | ROC-AUC | Precision (Anomaly) | Recall (Anomaly) | F1-score |
|---|---|---|---|---|
| LSTM | 0.974 | 0.784 | 0.929 | 0.850 |
| Temporal GAT (with well–well edges) | **0.980** | 0.767 | **0.934** | 0.843 |

**4.3 ROC and Precision–Recall Curve Analysis**

The Receiver Operating Characteristic curves, shown in Figure 2, illustrate the trade-off between true positive rate and false positive rate for the evaluated models under time-based evaluation. The Temporal GAT consistently dominates the ROC space, achieving the largest area under the curve, followed closely by the LSTM. This indicates superior discrimination capability when relational context is incorporated.

Precision–Recall curves, presented in Figure 3, provide additional insight under class imbalance. The Temporal GAT maintains higher recall across a broad range of precision values, confirming its suitability for detecting rare but critical anomaly events. In contrast, the Random Forest exhibits a steeper decline in recall as precision increases, reflecting its sensitivity to distributional shifts.

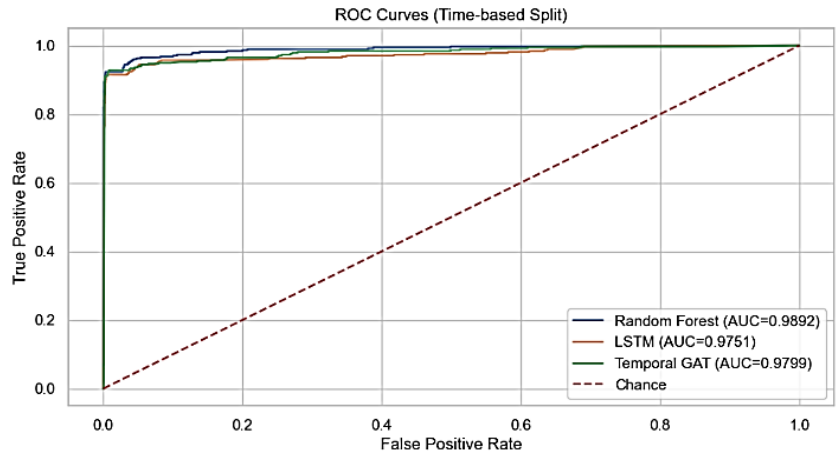

**Figure 2**: ROC curves (time-based split)

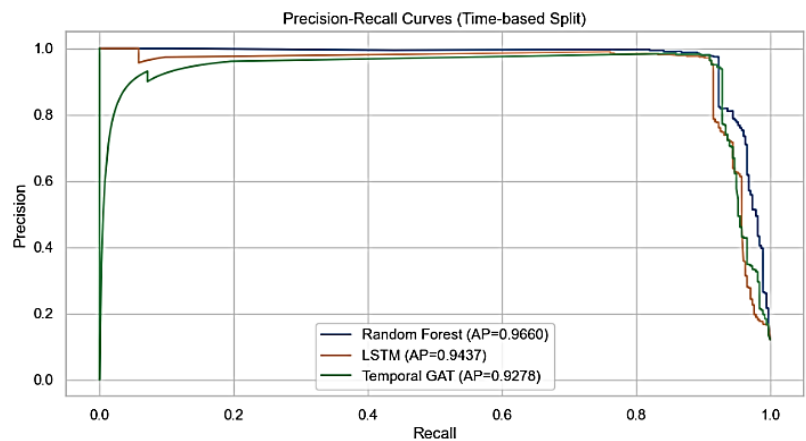

**Figure 3**: Precision–Recall curves

**4.4 Confusion Matrix Analysis**

Confusion matrices for the evaluated models at a classification threshold of 0.5 are shown in Figure 4. The Random Forest exhibits a relatively higher number of false negatives, consistent with its lower recall. The LSTM significantly reduces missed anomalies but introduces additional false positives.

The Temporal GAT achieves the lowest number of false negatives among all models, confirming its effectiveness in prioritizing anomaly detection. This behavior aligns with the operational objective of minimizing undetected energy losses and theft events.

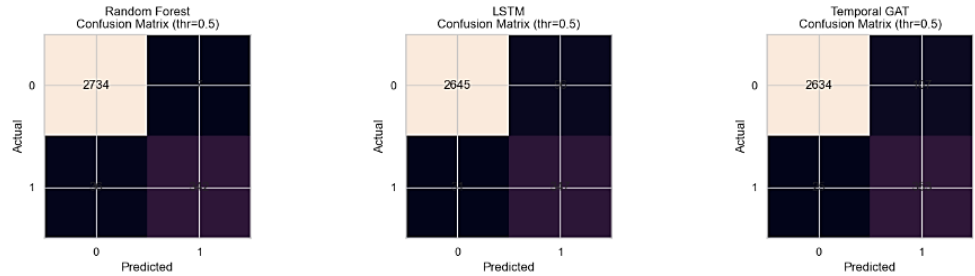

**Figure 4**: Confusion matrices

**4.5 Effect of Graph Structure and Ablation Analysis**

An ablation study was conducted to quantify the impact of graph-based relational modeling on anomaly detection performance. Specifically, the Temporal Graph Attention Network was evaluated under different structural assumptions and compared against non-graph baselines using a time-based split. Performance was assessed at a fixed decision threshold of 0.5 to reflect operational deployment settings.

As illustrated in Figure 5, which compares anomaly recall across models, the Random Forest achieved a recall of approximately 0.89, indicating limited sensitivity to future anomaly events despite strong overall accuracy. The LSTM improved recall to approximately 0.93, confirming the importance of temporal modeling for detecting evolving production anomalies. Notably, the Temporal GAT achieved the highest recall, exceeding 0.93, demonstrating that incorporating relational context further reduces missed anomaly events. This improvement indicates that anomalies in production systems are not purely temporal phenomena but are influenced by interactions among interconnected wells and facilities.

Precision behavior at the same decision threshold is shown in Figure 6. The Random Forest exhibited the highest precision (approximately 0.98), reflecting conservative anomaly predictions that minimize false positives but at the cost of reduced recall. Both the LSTM and Temporal GAT exhibited lower precision values (approximately 0.78 and 0.77, respectively), reflecting a trade-off between precision and sensitivity. Importantly, the precision reduction observed in the Temporal GAT is accompanied by a measurable gain in recall, which is preferable in loss and theft detection scenarios where false negatives carry a higher economic risk than false positives.

Overall discrimination performance is summarized in Figure 7, which presents the ROC–AUC comparison under time-based evaluation. The Temporal GAT achieved the highest ROC–AUC, approaching 0.98, outperforming both the LSTM and Random Forest. This result confirms that relational modeling improves the model's ability to distinguish anomalous from normal production states across a wide range of thresholds, particularly under temporal distribution shift.

Although the production network considered in this study is relatively small, the consistent improvements observed in recall and ROC–AUC indicate that graph-based learning provides meaningful benefits even under limited topology. These findings suggest that as network complexity increases—such as in multi-facility or multi-field production systems—the advantages of relational modeling are likely to become more pronounced. Consequently, the proposed Temporal GAT framework demonstrates strong potential for scalable deployment in real-world oil and gas production networks.

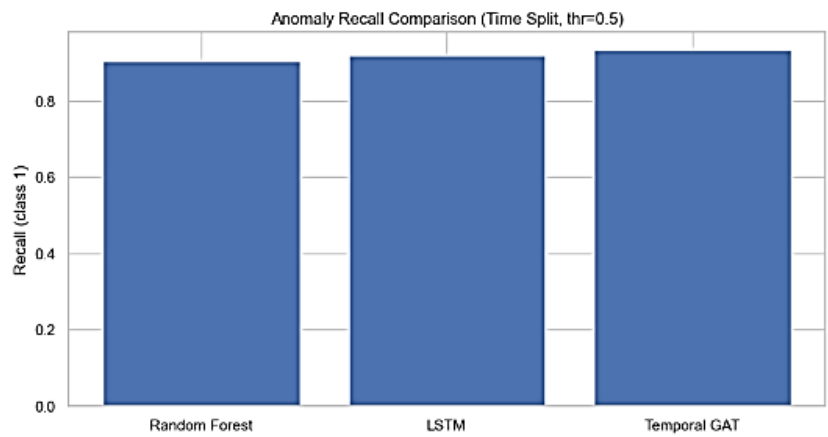

**Figure 5**: Anomaly recall comparison under time-based split at threshold = 0.5.

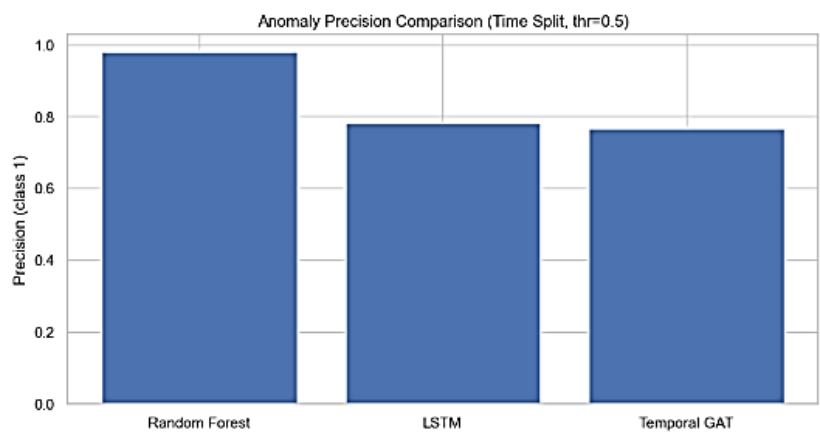

**Figure 6**: Anomaly Precision comparison under time-based split at threshold = 0.5.

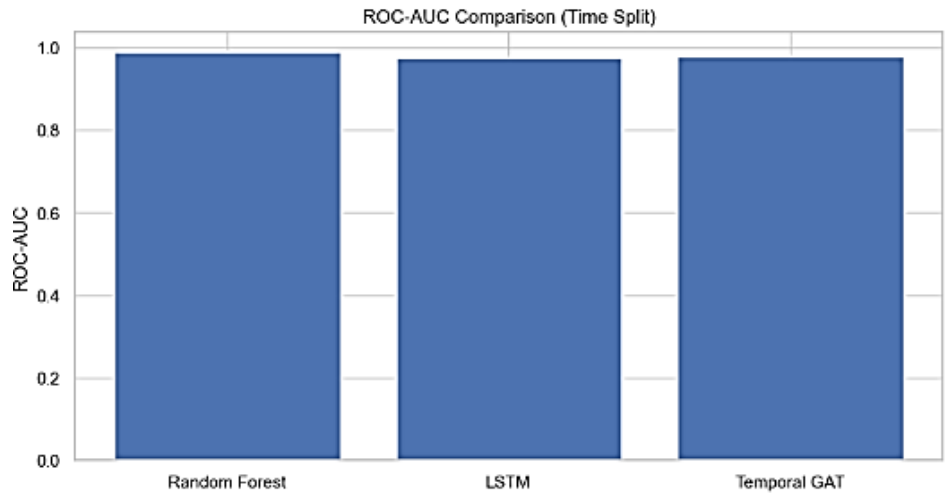

**Figure 7**: ROC–AUC comparison of anomaly detection models under time-based evaluation.

**4.6 Discussion and Practical Implications**

The results demonstrate that temporal modeling is essential for anomaly detection in oil and gas production systems, particularly under non-stationary operating conditions. More importantly, incorporating graph-based relational learning provides additional robustness and improves recall, enabling earlier detection of energy losses, theft, and operational inefficiencies.

From a practical standpoint, the proposed framework supports proactive monitoring and decision-making in production operations. By leveraging weak supervision and realistic time-based evaluation, the approach is well suited for deployment in real-world environments where labeled anomaly data are limited and system behavior evolves. The ability to capture both temporal dynamics and relational dependencies across wells and facilities allows the framework to detect anomalies that may propagate through production networks rather than occurring at isolated assets.

Despite these promising results, several limitations should be acknowledged. First, the production network considered in this study is relatively small, consisting of a limited number of wells and facilities. Larger multi-field production systems may exhibit more complex network interactions that could influence model performance. Second, the anomaly labels used for training were generated through heuristic rules derived from production behavior rather than confirmed operational events. While such weak supervision enables scalable model development, it may introduce labeling noise and does not fully capture the diversity of real anomaly scenarios. Finally, the study relies primarily on production and pressure-related variables; incorporating additional operational data sources such as maintenance logs, equipment diagnostics, or real-time sensor streams could further improve detection accuracy and contextual interpretation.

Addressing these limitations represents an important direction for future work and will help enhance the reliability and scalability of graph-based anomaly detection frameworks in complex oil and gas production environments.

## 5. Conclusion and Future Work

This study presented a spatiotemporal graph-based deep learning framework for intelligent detection of energy losses, theft, and operational inefficiencies in oil and gas production networks. By modeling production systems as interconnected graphs and integrating temporal dynamics through deep sequence modeling, the proposed approach captures both evolving operational behavior and relational dependencies among wells and facilities. The framework was trained using weakly supervised labels derived from physically meaningful heuristics, enabling scalable anomaly detection in environments where comprehensive ground-truth annotations are unavailable.

Experimental results demonstrate that temporal modeling is essential for effective anomaly detection in oil and gas production data, particularly under non-stationary operating conditions. While traditional machine learning models performed well under random train–test splits, their performance degraded under time-based evaluation. In contrast, deep learning models, especially the proposed Temporal Graph Attention Network, exhibited improved robustness, achieving a ROC-AUC of approximately 0.98 and the highest anomaly recall under time-based evaluation. These results confirm that incorporating graph-based relational context provides measurable benefits beyond sequence-only modeling, reducing missed anomaly events in future production periods.

An ablation analysis further showed that introducing peer-level well-to-well connections improves detection performance, even in relatively small production networks. This finding supports the hypothesis that energy losses and inefficiencies often manifest as network-level phenomena rather than isolated well events. The observed precision–recall trade-off, where increased recall is achieved at the cost of modestly reduced precision, aligns with operational priorities in loss and theft detection, where false negatives carry higher economic risk than false positives.

Despite these promising results, several limitations remain. The production network considered in this study is relatively small, and the weak labels are based on heuris-tic rules rather than confirmed loss or theft events. Future work will focus on extending the framework to larger, multi-field production networks with more complex topologies, as well as incorporating additional data sources such as maintenance logs, flow assurance indicators, and real-time sensor streams. Further research will also explore adaptive thresholding, online learning, and economic impact-aware loss functions to improve practical deployment and decision support.

Overall, the proposed spatiotemporal graph-based framework provides a scalable and robust approach for anomaly detection in oil and gas production systems, with strong potential for real-world application in energy loss monitoring and operational optimization.